\pdfoutput=1

\documentclass[11pt]{article}

\usepackage[preprint]{acl}

\usepackage{times}
\usepackage{latexsym}

\usepackage[T1]{fontenc}

\usepackage[utf8]{inputenc}

\usepackage{microtype}

\usepackage{inconsolata}

\usepackage{graphicx}

\title{ 
Exploring NLP Benchmarks in an Extremely Low-Resource Setting
}

\author{
  Ulin Nuha\thanks{Work done while the author was a PhD student at NKUST.} \\
  National Kaohsiung University \\
  of Science and Technology \\
  \texttt{ulinnuha.id@icloud.com}
  \And
  Adam Jatowt \\
  University of Innsbruck \\
  \texttt{adam.jatowt@uibk.ac.at}
}

\begin{document}
\maketitle
\begin{abstract}
The effectiveness of Large Language Models (LLMs) diminishes for extremely low-resource languages, such as indigenous languages, primarily due to the lack of labeled data. Despite growing interest, the availability of high-quality natural language processing (NLP) datasets for these languages remains limited, making it difficult to develop robust language technologies. This paper addresses such gap by focusing on Ladin, an endangered Romance language, specifically targeting the Val Badia variant. Leveraging a small set of parallel Ladin–Italian sentence pairs, we create synthetic datasets for sentiment analysis and multiple-choice question answering (MCQA) by translating monolingual Italian data. To ensure linguistic quality and reliability, we apply rigorous filtering and back-translation procedures in our method. We further demonstrate that incorporating these synthetic datasets into machine translation training leads to substantial improvements over existing Italian–Ladin translation baselines. Our contributions include the first publicly available sentiment analysis and MCQA datasets for Ladin, establishing foundational resources that can support broader NLP research and downstream applications for this underrepresented language.
\end{abstract}

\section{Introduction}
\begin{figure}[!t]
  \centering
  \includegraphics[width=2.35in]{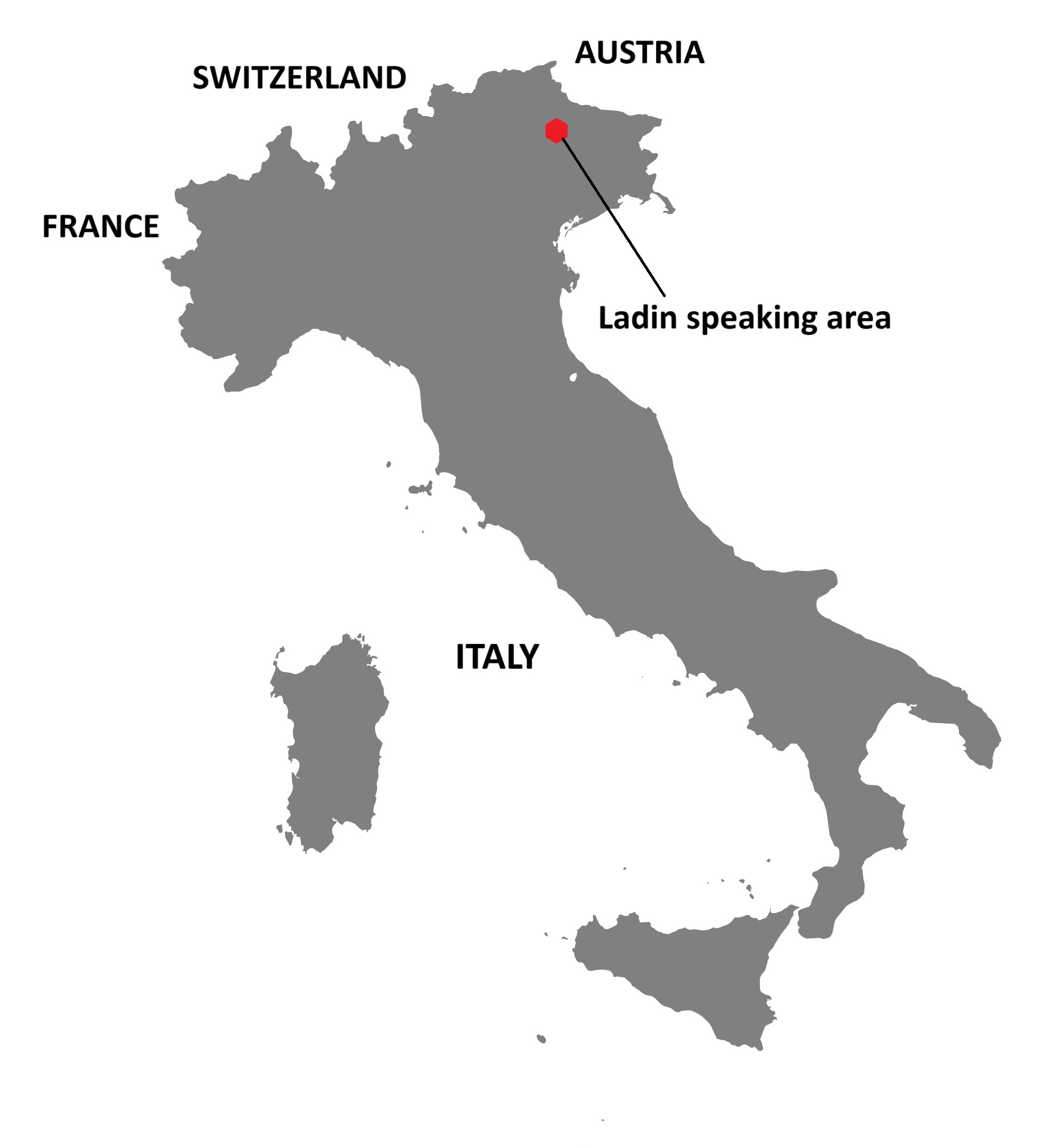}
  \caption{A map highlighting the Ladin speaking region in South Tyrol, Northern Italy.}
  \label{fig:LadinArea}
\end{figure}

Large language models (LLMs) have garnered significant attention in diverse audiences \citep{Yuxia:2024, Gambardella:2024} due to their ability to effectively perform natural language tasks with only a few input-output examples \citep{Cheng:24}, while also eliminating the need for gradient updates of the model \citep{Nguyen:2024}. 
LLMs achieve remarkable performance through pre-training on large corpora, but their reliance on high-resource languages (HRLs) limits their effectiveness for low-resource languages (LRLs) \citep{Pham:2024}, especially extreme ones \citep{purason2024llms}. Efforts to address this limitation include various techniques such as in-context learning \citep{Cahyawijaya:2024} and fine-tuning \citep{Alabi:2022, Su2024Unlocking} to transfer LLM capabilities to LRLs. Additionally, some studies \citep{yong2024lexc, morim2024, tran24irish} have focused on developing translation systems between LRLs and HRLs to address the lack of language technology resources. The limited advancement of natural language processing (NLP) in LRLs stems largely from the disproportionate focus of research on improving language models for HRLs, often at the expense of developing tools and resources for low-resource counterparts \citep{zhang24llm}. Moreover, the scarcity of NLP datasets for extremely low-resource languages (ELRLs) constitutes a major barrier to progress in this area, highlighting the need for more inclusive AI technologies to support marginalized languages.

This paper advances NLP research for Ladin language, an Indigenous and extremely low-resource language of Northern Italy (see Fig.~\ref{fig:LadinArea}). 
Despite progress in NLP for LRLs, Ladin’s dialectal diversity and limited digitized corpora pose unique challenges. Building on the prior work \citep{frontull24}, which established a machine translation (MT) benchmark between Italian and Ladin (Val Badia variant), we further improve the translation performance for Ladin.

Ladin is a Rhaeto-Romance language spoken by about 30,000 people in South Tyrol, Northern Italy. It comprises five dialects (Val Badia, Fascia, Anpezo, Fodom, and Gherdëina) which vary significantly in morpho-syntactic and orthographic conventions. Although standardized efforts such as Ladin Dolomitan exist, it is not officially recognized and is used only to a limited extent, while speakers report minimal familiarity with Ladin varieties beyond their own \citep{connor2023ladin}. Linguistic conventions have also changed over time within dialects. This linguistic diversity presents major challenges for machine translation and resource creation. Publicly available resources are extremely limited, with only a few lexicons, dictionaries, and minor corpora. There is a major source of monolingual Ladin data in \textit{La Usc di Ladins}\footnote{https://www.lausc.it/}, a newspaper published in five variants and digitally archived since 2012, but this resource remains underutilized due to the lack of alignment and limited annotation. Our work focuses on the Val Badia variant of Ladin
, due to the availability of the largest accessible dataset. Previous work on MT for Ladin's Fascia variant exists \citep{FasciaLadin}; however, the parallel dataset used in that work consists of only 1,135 sentences. 

Unlike prior studies, our work also extends beyond MT for Ladin to include sentiment analysis (SA) and multiple-choice question answering (MCQA), establishing the first datasets for these tasks in Ladin. First, we compare MT approaches, including few-shot learning and fine-tuning, using available Ladin–Italian sentence pairs. Subsequently, we construct high-quality synthetic Ladin datasets for SA and MCQA through the translation from monolingual Italian data, applying rigorous filtering to ensure high quality. These datasets not only contribute to SA and MCQA but also enhance MT. To assess this, we incorporate our synthetic dataset into existing Italian–Ladin translation benchmarks and evaluate its impact.

In summary, the primary contributions of our work can be outlined as follows:
\begin{itemize}
    \item We conduct a comparative study on translation between Italian and Ladin, focusing on the Val Badia variant. Our model significantly outperforms the current benchmark, demonstrating the effectiveness of our approach for this extremely low-resource language (LRL).
    
    \item We construct a high-quality synthetic dataset of Ladin–Italian sentence pairs, referred to as \textit{SD\textsubscript{Lad--Ita}}, derived from monolingual Italian data using a language model and rigorous filtering techniques.
    
    \item We showcase the utility of the synthetic dataset beyond MT by applying it to additional downstream NLP tasks, including SA (text classification) and MCQA, establishing the first such datasets in Ladin.
\end{itemize}

By fostering NLP research on Ladin, we aim to support the preservation and accessibility of this critically underrepresented language—benefiting education, digital communication, and cultural documentation for Indigenous communities.

\section{Background and Related Work}

NLP for ELRLs has garnered increasing attention from researchers, largely due to the enduring challenges associated with data scarcity. These languages typically possess fewer than 0.1 million available parallel sentence pairs \citep{Ranathunga23}, which are often insufficient for effectively training neural machine translation (NMT) models \citep{rudra:19}. For instance, Ladin has fewer than 100 thousand parallel sentences available for MT \citep{frontull24}, while resources for other NLP tasks, such as text classification and question answering, remain non-existent. 

Ladin and Italian are both Romance languages with shared Latin roots, resulting in potential similarities in lexical and syntactic aspects. This linguistic proximity facilitates cross-lingual transfer, allowing multilingual models fine-tuned on Italian to generalize moderately well to Ladin. However, key divergences remain due to Ladin’s unique phonological and morphological traits \citep{LucaM}. These differences are further shaped by historical geographic isolation and varying degrees of contact with dominant surrounding languages, particularly regional varieties of German and Italian. Most Ladin speakers in Italy are bi- or tri-lingual, using Ladin in private domains and Italian, German, or both in public settings, as neighboring communities typically lack comprehension of Ladin \citep{SilviaLadin}. This contact has led to regional variation. For instance, Gherdëina and Val Badia exhibit stronger German influence, while southern valleys show more Italian features. Val Badia is often perceived as having the “purest” Ladin, though this reflects relative rather than absolute isolation from external influence. These dynamics pose several challenges for translation. The lack of a commonly adopted standard variety necessitates dialect-specific translation strategies, complicating mutual intelligibility and the development of shared resources. Furthermore, long-standing language contact has introduced borrowings and calques, making it difficult to isolate core Ladin structures.

To address the scarcity of resources in ELRLs, recent efforts have focused on leveraging transfer learning techniques, particularly through the use of large language models (LLMs) \citep{Pham:2024, lim:24}. However, success remains limited in such settings \citep{tran24irish}, where extreme data sparsity poses challenges for effective domain adaptation and language alignment. In addition, a key aspect of transfer learning with LLMs involves leveraging token similarity and cross-region similarity to better capture shared cultural and linguistic features \citep{SinaBN}. \citet{RAG-LLM} proposed a MT framework that integrates Retrieval-Augmented Generation (RAG) with LLMs to address these challenges, whereas \citet{DPO-LLM} explored the use of LLMs in combination with Direct Preference Optimization (DPO) to enhance translation quality. 

Despite promising results, the above-mentioned methods incur high computational costs. Multimodal approaches, which incorporate additional modalities such as visual context to enhance translation \citep{rajpoot:24, sami:dcu, hatami:24}, require supplementary datasets that are often unavailable for these languages and face considerable computational and complexity challenges. Then, Rule-based machine translation (RBMT), exemplified by Apertium \citep{garcia:24, sanchez:24}, offers a less resource-intensive alternative, with lower computational requirements compared to NMT systems. However, RBMT systems are labor-intensive, requiring extensive manual effort to create and maintain linguistic rules, with scalability further hindered by the challenge of ensuring rule self-consistency \citep{liu-multil:24}. 

Another approach to supporting technologies for LRLs is the development of benchmarks. For example, \citet{urbizu} introduced BasqueGLUE, a benchmark covering multiple NLP tasks for the Basque language.
Similarly, \citet{FarsInstruct} presented FarsInstruct, a large-scale dataset developed to strengthen LLMs’ capacity to follow instructions in Persian, a globally low-represented language. However, recent work in language revitalization and human-centered NLP shows the importance of going beyond performance metrics in developing technologies for endangered languages \citep{StevenBird}. Effective systems should align with long-term community goals, such as intergenerational transmission and cultural continuity.

In line with this perspective, our work presents a comparative study of MT approaches, including instruction learning and fine-tuning \citep{zhang:mt23}, using several language models such as NLLB and GPT-4. 
To evaluate translation quality, we use both statistical and semantical metrics, particularly for synthetic data. While prior research has made progress in NLP for LRLs, substantial gaps remain for ELRLs such as Ladin. This study addresses these gaps by leveraging data augmentation techniques and introducing benchmark datasets for MT, text classification, and question answering tasks.

\section{Experiments}

\begin{figure*}[t]
  \centering
  \includegraphics[width=5.3in]{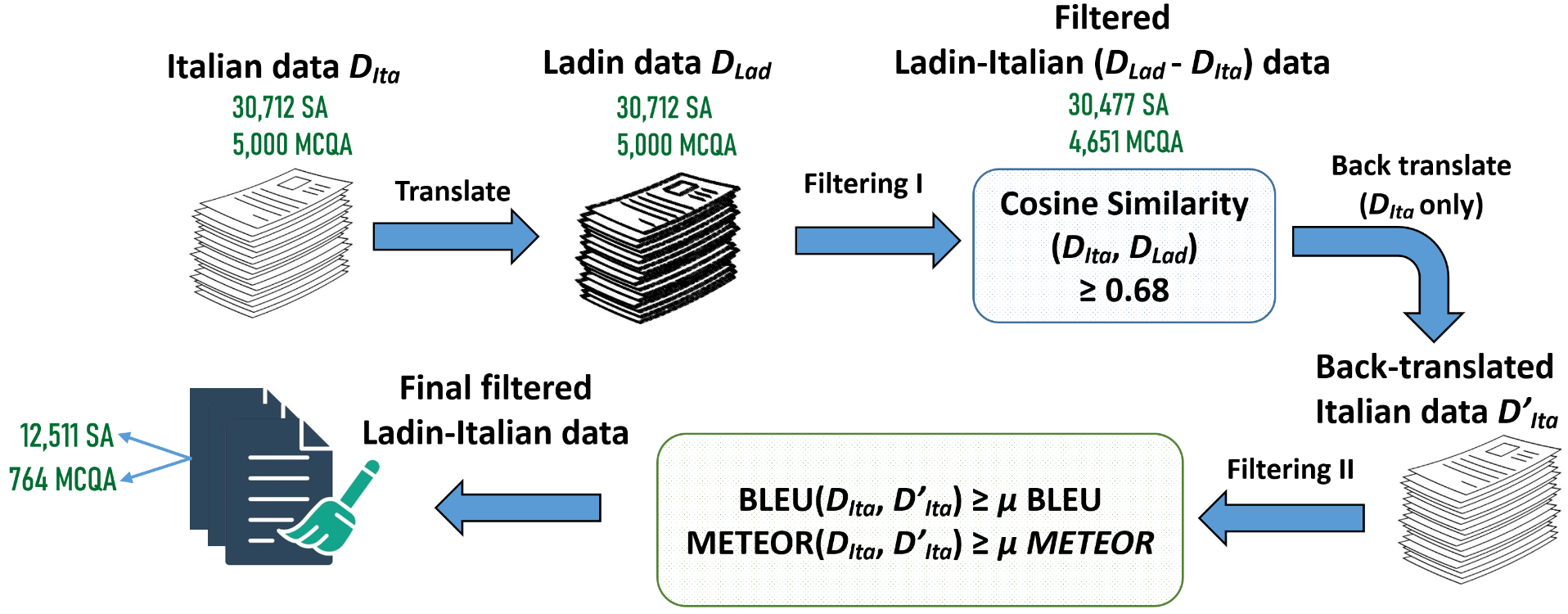}
  \caption{Synthetic data generation process. Initially, we translate the Italian data  \( D\textsubscript{\textit{Ita}}\) of SA and MCQA into Ladin. The Ladin translation  \(D\textsubscript{\textit{Lad}}\) of SA and MCQA is then filtered using Filtering I. Next, the filtered \(D\textsubscript{\textit{Lad}}\) is back-translated into Italian \( D^{\prime}\textsubscript{\textit{Ita}}\), followed by Filtering II to obtain the high-quality parallel paired datasets of SA and MCQA in Ladin and Italian.}
  \label{fig:experiments}
\end{figure*}


\subsection{Data sources}
In this section, we provide an overview of the data sources used in our study. We prepared datasets for three tasks: MT, SA, and MCQA. These resources were collected from publicly available datasets. Figure~\ref{fig:experiments} illustrates the synthetic paired Ladin-Italian data generation process. As the creation of the SA and MCQA datasets depends on the availability of a translation model, we first constructed the MT dataset. Once the translation model was established, we proceeded with the generation of the SA and MCQA datasets accordingly.
\noindent\textbf{Machine translation.} The parallel Italian–Ladin dataset used for both training and testing in our MT experiments was initially derived from prior work \citep{frontull24}. The authentic Italian-Ladin training dataset \textit{AD\textsubscript{Ita\_Lad}} comprises 18,139 sentence pairs, carefully crafted as basic and concise examples to illustrate the use of specific words and phrases. 
The average sentence length in \textit{AD\textsubscript{Ita\_Lad}} is 23.43 characters for Ladin and 25.69 characters for Italian, while the average number of words per sentence is 5.02 for Ladin and 4.36 for Italian. Following that work, the testing dataset, \textit{\( T = \{t\textsubscript{1}, t\textsubscript{2}, t\textsubscript{3}\} \)}, consists of three subsets: \( t_1 \) (424 sentences) focuses on formal and legal terminology; \( t_2 \) (833 sentences) blends stylistic and lexical elements, reflecting regional cultural and historical contexts; \( t_3 \) (1,563 sentences) includes narrative prose, dialogue, and idiomatic expressions, offering diverse challenges. For Italian in the testing dataset, the average word counts for \( t_1 \), \( t_2 \), and \( t_3 \) are 21.25, 26.65, and 15.61, respectively. In comparison, the corresponding averages for \( t_1 \), \( t_2 \), and \( t_3 \) in Ladin are 23.58, 24.52, and 13.05, respectively.

\noindent\textbf{Sentiment analysis.} We construct datasets for NLP tasks in Ladin by leveraging labeled monolingual Italian resources, including SA and MCQA datasets. We use the monolingual Italian SA dataset \textit{D\textsubscript{Ita\_SA}} from \citet{desole2025} for the SA task. \textit{D\textsubscript{Ita\_SA}} comprises abundant Tripadvisor reviews labeled as positive or negative. The initial dataset contains 41,077 entries. To mitigate the impact of excessively long reviews, we filter the dataset to retain only those entries with word counts up to the third quartile (Q3), resulting in a maximum review length of 138 words. This screening results in a reduced dataset size of 30,712 entries. Given that the raw dataset, collected via web scraping, contains grammatical errors, we employ GPT-4 to correct these while preserving the original semantic content. Following grammar correction, the average number of words per sentence in \textit{D\textsubscript{Ita\_SA}} is 70.44.

\noindent\textbf{MCQA.} Finally, for the MCQA task, we use the monolingual Italian MCQA dataset \textit{D\textsubscript{Ita\_MCQA}} from \citet{multi_it24}. \textit{D\textsubscript{Ita\_MCQA}} comprises over 5,000 manually crafted questions covering a wide range of topics. This MCQA dataset contains questions with 2, 3, 4, 5, or 6 answer choices. However, we exclude questions with 2 and 6 options due to their very limited number.

\subsection{Language model settings for machine translation}
We now discuss the process of building a MT model using existing Italian–Ladin parallel data, which serves as a prerequisite for the subsequent synthetic data generation process. To facilitate translation between Italian and Ladin, we employ both LLMs and sequence-to-sequence (seq2seq) models. For LLM-based approaches, we utilize LLaMA 3.1 (8B and 70B variants) and GPT-4o. The LLaMA models are evaluated in two configurations: few-shot learning (FSL) and supervised fine-tuning (FT). Appendix ~\ref{sec:fsl} provides an example prompt used for the FSL approach. To perform the FSL approach, we utilize the \textit{LLaMA-v3.1-8b-instruct} and \textit{LLaMA-v3.1-70b-instruct} models via the DeepInfra API. Similarly, the FSL approach with GPT-4o is conducted using the OpenAI\footnote{https://deepinfra.com/} API. For the FT approach using LLaMA, we leverage the Together AI\footnote{https://www.together.ai/}  fine-tuning API with the LoRA adapter configuration (\textit{r} = 32), employing a batch size of 8 and 3 epochs.

For the seq2seq models, we employ MBART-large-50 and NLLB-200-1.3B, both of which are encoder-decoder architectures specifically designed for multilingual MT. The fine-tuning models of NLLB-1.3B and MBART-large-50 are conducted on an NVIDIA RTX A6000 GPU with a batch size of 8 and over 7 epochs. To accommodate Ladin, an unseen language for both seq2seq models, we introduce a Ladin-specific language tag in the tokenizer. This modification aids in the identification and processing of Ladin texts during translation tasks, facilitating better handling of the language.

Subsequently, we evaluate translation performance using standard metrics: Sacre BLEU, ROUGE, and chrF++, providing a comprehensive assessment of translation quality.

\subsection{Synthetic data creation}

\begin{table*}
  \centering
  \small
  \begin{tabular}{llcccccc}
    \hline
    \textbf{Training Data}     & \textbf{Model}      
    & \multicolumn{2}{c}{\textbf{BLEU}} 
    & \multicolumn{2}{c}{\textbf{Rouge}} 
    & \multicolumn{2}{c}{\textbf{chrF++}} \\
    \cline{3-8}
    &
    & \textbf{Ita→Lad} & \textbf{Lad→Ita} 
    & \textbf{Ita→Lad} & \textbf{Lad→Ita} 
    & \textbf{Ita→Lad} & \textbf{Lad→Ita} \\
    \hline
   \textit{AD\textsubscript{Ita\_Lad}} & FSL-Llama 3.1 8B         & 2.87             & 12.09            & 18.48            & 36.79            & 23.43            & 36.20             \\
   & FSL-Llama 3.1 70B        & 6.35             & 21.97            & 31.81            & 51.09            & 31.09            & 47.26            \\
   & FSL-GPT-4o               & 4.27             & \underline{22.84}            & 25.86            & \underline{52.28}            & 27.47            & 48.23            \\
   & FT-Llama 3.1 8B          & 10.71            & 15.28            & 41.56            & 43.78            & 35.23            & 42.04            \\
   &  FT-Llama 3.1 70B         & 6.95             & 22.48            & 32.97            & 50.77            & 31.96            & 47.96            \\
    & Mbart-large-50           & 10.37            & 10.91            & 44.14            & 41.41            & 41.73            & 43.92            \\
   & FT-NLLB 1.3 B               & \underline{17.76}            & 21.41            & \underline{52.81}            & 49.91            & \underline{44.60 }           & \underline{48.40}            \\
   & Benchmark (LLM)                & 3.51             & 19.44            & -                & -                & 25.54            & 44.69            \\
  
  \hline
  \textit{AD\textsubscript{Ita\_Lad}} + \textit{SD\textsubscript{Ita\_Lad}} 
  & FT-NLLB 1.3 B               & 18.30            & \textbf{24.50}            & \textbf{53.66}            & \textbf{52.64}            & \textbf{44.62}           & \textbf{50.76}            
  \\
  \rule{0pt}{5.2ex} 
  \shortstack[l]{\textit{AD\textsubscript{Ita\_Lad}} +\\monolingual data} 
  & \shortstack[l]{Benchmark\\(RBMT)}                 
  & \textbf{18.97}             & 19.32            & -                & -                & 44.13            & 46.69            \\
    \hline
  \end{tabular}
  \caption{\label{translation-performance}
     Comparison of different models based on average translation performance metrics on the test set \( T \). The table is divided into two sections: the upper section reports results using only authentic training data \textit{AD\textsubscript{Ita\_Lad}}, while the lower section presents results with additional synthetic data.}

\end{table*}

We aim to create datasets for NLP tasks in the Ladin language by translating \textit{D\textsubscript{Ita\_SA}} and \textit{D\textsubscript{Ita\_MCQA}} from Italian data \textit{D\textsubscript{Ita}} to Ladin data \textit{D\textsubscript{Lad}}. These translated datasets serve multiple purposes, including translation (Italian–Ladin pairs), text classification, and question answering tasks. To ensure high-quality translation, we first select the best-performing MT model—identified in the previous section based on its performance on a held-out testing data \( T \), and use this model to translate \textit{D\textsubscript{Ita}} into \textit{D\textsubscript{Lad}}. The creation of this synthetic dataset shown in Figure~\ref{fig:experiments} involves several key steps, as described below.

\noindent\textbf{Italian-to-Ladin translation.} We begin by translating the labeled monolingual Italian datasets \textit{D\textsubscript{Ita}} of SA and MCQA tasks into Ladin, resulting in  \textit{D\textsubscript{Lad}}. This translation process is carried out using the best-performing MT model identified in our evaluation, ensuring the highest possible quality of the synthetic Ladin data.

\noindent\textbf{Filtering I.} To ensure the semantic quality of the Italian → Ladin translations, we apply a filtering step using similarity scores computed by the Language-Agnostic BERT Sentence Embedding (LaBSE) model \citep{labse:22}. This step evaluates the alignment between each translated Ladin sentence and its original Italian counterpart. Specifically, we retain only those sentence pairs with a cosine similarity score  \( c \geq 0.68 \), which corresponds to the average similarity observed in the aligned dataset \textit{AD\textsubscript{Ita\_Lad}}. This threshold helps eliminate semantically inconsistent translations while preserving high-quality aligned pairs for downstream tasks.

\noindent\textbf{Back-translation.} We employ back-translation to further refine the synthetic dataset prior to conducting a second round of filtering. Specifically, the Ladin translations \( D\textsubscript{\textit{Lad}} \) are retranslated into Italian, yielding \( D^{\prime}\textsubscript{\textit{Ita}} \), which are then compared against the original Italian datasets \( D\textsubscript{Ita} \). This step helps identify and discard translations that deviate significantly from the original semantics, thereby improving data quality.

\noindent\textbf{Filtering II.} To finalize the synthetic dataset, we apply a second filtering step based on automatic evaluation metrics. Specifically, we compute the Sacre BLEU and METEOR scores between the original Italian data (\( D\textsubscript{\textit{Ita}} \) and the back-translated Italian data \( D^{\prime}\textsubscript{\textit{Ita}} \)). A translation instance is retained if  
\( BLEU \)(\( D\textsubscript{\textit{Ita}} \), \( D^{\prime}\textsubscript{\textit{Ita}} \)) \(\geq \mu{BLEU} \) 
and 
\( METEOR \)(\( D\textsubscript{\textit{Ita}} \), \( D^{\prime}\textsubscript{\textit{Ita}} \)) \(\geq \mu{METEOR} \).
where \( \mu{BLEU} \) and \( \mu{METEOR} \) represent the average scores computed across all translation pairs. These threshold values ensure semantic fidelity and fluency by eliminating noise and inaccuracy, thus improving the quality of the synthetic dataset.

\subsection{Synthetic data evaluation}

To evaluate the quality and impact of the Italian–Ladin synthetic parallel dataset \textit{SD\textsubscript{Ita\_Lad}} = {(\textit{D\textsubscript{Ita\_SA}}, \textit{D\textsubscript{Lad\_SA}}), (\textit{D\textsubscript{Ita\_MCQA}}, \textit{D\textsubscript{Lad\_MCQA}})}, we combine it with the authentic dataset \textit{AD\textsubscript{Ita\_Lad}} and assess performance on the testing data \( T \). This setup allows us to measure the effect of synthetic data augmentation on translation quality. To assess lexical and syntactic adequacy of the synthetic translations, we construct a manually translated gold dataset \( GD \)
, containing 50 examples each from \textit{D\textsubscript{Ita\_SA}} and \textit{D\textsubscript{Ita\_MCQA}}, translated into Ladin by a native speaker of Italian and Ladin. We compute cosine similarity between the synthetic and gold translations using sentence embeddings, providing a measure of semantic alignment.

\section{Results and Discussion}
In this section, we report the results of MT using the existing Italian–Ladin parallel data. We then present the evaluation of the synthetic Ladin datasets for SA and MCQA, including both quantitative metrics and qualitative analysis.

\begin{table*}[!t]
  \centering
 \small
  \begin{tabular}{llccccccccc}
    \hline
    \textbf{Training Data}     & \textbf{Model}    
    & \multicolumn{3}{c}{\textbf{BLEU}} & \multicolumn{3}{c}{\textbf{Rouge}} 
    & \multicolumn{3}{c}{\textbf{chrF++}} \\
     \cline{3-11} &
                            & \textbf{\boldmath\( t_1 \)} & \textbf{\boldmath\( t_2 \)} & \textbf{\boldmath\( t_3 \)} &
                            \textbf{\boldmath\( t_1 \)} & \textbf{\boldmath\( t_2 \)} & \textbf{\boldmath\( t_3 \)} & 
                            \textbf{\boldmath\( t_1 \)} & \textbf{\boldmath\( t_2 \)} & \textbf{\boldmath\( t_3 \)} \\
    \hline
    \textit{AD\textsubscript{Ita\_Lad}} & FSL-Llama 3.1 8B    
    & 5.12  & 2.45  & 1.03    & 22.87  & 19.00 & 13.56   & 27.36   
    & 26.00 & 16.92  \\
     & FSL-Llama 3.1 70B     
    & 8.43  & 7.66  & 2.96    & 35.49  & 33.85 & 26.09   & 35.05   
    & 35.05 & 23.17  \\
     & FSL-GPT-4o     
    & 6.62  & 4.15  & 2.03    & 30.18  & 26.04 & 21.36   & 31.90   
    & 30.08 & 20.42  \\
     & FT-Llama 3.1 8B     
    & 12.70  & 9.79  & 9.64    & 43.20  & 40.03 & 41.44   & 38.11   
    & 36.60 & 30.98  \\
     & FT-Llama 3.1 70B     
    & 9.41  & 7.75  & 3.70     & 36.83  & 34.60 & 27.47   & 36.03   
    & 35.87 & 23.99  \\
     & Mbart-large-50     
    & 11.28  & 11.34  & 11.93    & 43.56  & 42.75 & 46.12   & 39.8   
    & 38.66 & 35.23  \\
     & NLLB 1.3 B     
    & \underline{18.54}  & \underline{16.85}  & \underline{17.88}    & \textbf{51.59} & \underline{51.55} & \underline{55.29}   & \textbf{46.24}
    & \underline{45.98} & \underline{41.58}  \\
     & Benchmark (LLM)     
    & 5.54  & 3.84  & 1.16    & -  & - & -   & 29.03   
    & 28.98 & 18.60  \\
\hline
  \textit{AD\textsubscript{Ita\_Lad}} + \textit{SD\textsubscript{Ita\_Lad}} 
  & FT-NLLB 1.3 B             
  & 18.30  & 17.69  & \textbf{18.29}    & 51.40  & \textbf{52.92} & \textbf{56.65}   & 45.44  
  & 46.28 & \textbf{42.15} 
  \\
  \rule{0pt}{5.2ex} 
  \shortstack[l]{\textit{AD\textsubscript{Ita\_Lad}} +\\monolingual data} 
  & \shortstack[l]{Benchmark\\(RBMT)}                
  & \textbf{20.93}  & \textbf{19.32}  & 16.65    & -  & - & -   & \textbf{47.65} & \textbf{46.58} & 38.16  \\\\      
    \hline
  \end{tabular}
  \caption{\label{add-trans_to_ladin}
     Translation performance of various models for Italian → Ladin across three test subsets \( t_1 \), \( t_2 \), and \( t_3 \). The table is divided into two sections: the upper section reports results using only authentic training data \textit{AD\textsubscript{Ita\_Lad}}, while the lower section presents results with additional synthetic data.
  }
\end{table*}

\begin{table*}[!t]
  \centering
  \small
  \begin{tabular}{llccccccccc}
    \hline
    \textbf{Training Data}     & \textbf{Model}    
    & \multicolumn{3}{c}{\textbf{BLEU}} & \multicolumn{3}{c}{\textbf{Rouge}} 
    & \multicolumn{3}{c}{\textbf{chrF++}} \\
     \cline{3-11} &
                            & \textbf{\boldmath\( t_1 \)} & \textbf{\boldmath\( t_2 \)} & \textbf{\boldmath\( t_3 \)} &
                            \textbf{\boldmath\( t_1 \)} & \textbf{\boldmath\( t_2 \)} & \textbf{\boldmath\( t_3 \)} & 
                            \textbf{\boldmath\( t_1 \)} & \textbf{\boldmath\( t_2 \)} & \textbf{\boldmath\( t_3 \)} \\
    \hline
    \textit{AD\textsubscript{Ita\_Lad}} & FSL-Llama 3.1 8B    
    & 19.25  & 12.74  & 4.27    & 50.09  & 38.71 & 21.56   & 46.30   
    & 38.89 & 23.41  \\
     & FSL-Llama 3.1 70B     
    & 30.07  & 21.55  & 14.29    & 62.16  & 51.60 & 39.51   & 56.22   
    & 48.66 & 36.91  \\
     & FSL-GPT-4o     
    & 29.23  & 23.38  & 15.90    & 61.39  & \textbf{53.37} & 42.07   & 55.73   
    & 50.33 & 38.64  \\
     & FT-Llama 3.1 8B     
    & 20.53  & 17.25  & 8.05    & 51.86  & 44.85 & 34.62   & 48.36   
    & 44.11 & 33.66 \\
    &  FT-Llama 3.1 70B     
    & \textbf{30.78}  & \textbf{23.65}  & 13.02     & \textbf{61.91}  & 52.10 & 38.29   & \textbf{57.23}   
    & \underline{50.53} & 36.12  \\
     & Mbart-large-50     
    & 11.95  & 12.81  & 11.84    & 40.15  & 42.97 & 41.12   & 41.11   
    & 42.15 & 36.81  \\
    &  NLLB 1.3 B     
    & 27.15  & 19.77  & \underline{17.31}    & 55.54 & 48.27 & \underline{46.03}   & 54.45
    & 49.66 & \underline{41.09}  \\
    &  Benchmark (LLM)     
    & 26.77  & 21.17  & 10.37    & -  & - & -   & 53.20   
    & 48.52 & 32.36  \\
\hline
  \textit{AD\textsubscript{Ita\_Lad}} + \textit{SD\textsubscript{Ita\_Lad}} 
  & FT-NLLB 1.3 B             
  & \underline{30.46}  & \underline{22.71}  & \textbf{20.33}    & 58.11  & 50.71 & \textbf{49.11}   & \underline{56.95}   
  & \textbf{51.49} & \textbf{43.83} 
  \\
  \rule{0pt}{5.2ex} 
  \shortstack[l]{\textit{AD\textsubscript{Ita\_Lad}} +\\monolingual data} 
  & \shortstack[l]{Benchmark\\(RBMT)}                
  & 21.36  & 20.27  & 16.34    & -  & - & -   & 50.24   
    & 49.08 & 40.76  \\\\    
    \hline
  \end{tabular}
  \caption{\label{add-trans_to_italian}
     Translation performance of various models for Ladin → Italian across three test subsets \( t_1 \), \( t_2 \), and \( t_3 \). The table is divided into two sections: the upper section reports results using only authentic training data \textit{AD\textsubscript{Ita\_Lad}}, while the lower section presents results with additional synthetic data.
  }
\end{table*}

\subsection{Machine translation of Italian and Ladin}
Table~\ref{translation-performance} presents the translation performance results for both Italian → Ladin and Ladin → Italian across diverse language model configurations. For experiments using \textit{AD\textsubscript{Ita\_Lad}} as the training data, we report results from the benchmark approach proposed by \citet{frontull24}, which employs the LLM \verb|GPT-3.5-turbo-0125|, for comparison purposes. 
While Table~\ref{translation-performance} presents the overall average performance metrics of each translation model evaluated on the testing data \( T \), Tables~\ref{add-trans_to_ladin} and~\ref{add-trans_to_italian} offer a more fine-grained analysis. They report detailed metric scores for the individual test subsets \( t_1 \), \( t_2 \), and \( t_3 \), allowing for a closer examination of model behavior across different data segments.

In both translation directions in Table~\ref{translation-performance}, the fine-tuned NLLB (FT-NLLB) model achieves significant performance across evaluation metrics. Specifically, the model performs best on the translation from Italian → Ladin, with a BLEU score approaching 18, indicating relatively good translation accuracy for extremely low-resource settings. Although the Ladin → Italian translation performance of the FT-NLLB model slightly lags behind that of the FT-LLaMA 3.1 70B model, the FT-NLLB model achieves the highest chrF++ scores in both translation directions, indicating superior overall translation quality. These findings underscore a persistent challenge in extremely low-resource NLP. Despite the impressive capabilities of LLMs, their performance often suffers in extremely low-resource scenarios due to training biases toward HRLs with abundant data. Consequently, the Ladin → Italian translation achieves higher BLEU scores, reflecting the model’s stronger proficiency in HRLs. 

Although Ladin and Italian both belong to the Romance language family, translation into Ladin remains challenging due to its status as the extremely low-resource language, including unique vocabulary, dialectal variations, and a lack of direct equivalence to Italian. In contrast, the NLLB model demonstrates relatively robust performance, likely benefiting from its multilingual training across 200+ languages, including over 150 LRLs \citep{nllb2022}. Notably, the NLLB model was trained on Friulian, a closely related Raeto-Romance language \citep{UNESCO}, which may contribute to its improved generalization on Ladin.

Additionally, we evaluate the suitability and effectiveness of the NLLB tokenizer for processing Ladin text. Our analysis reveals an average tokens-per-word ratio of 1.50 for Ladin, compared to 1.39 for Italian. This modest overhead indicates that the NLLB tokenizer segments Ladin reasonably well, despite the language's extremely low-resource status, suggesting its viability for downstream translation tasks.

\subsection{Qualitative error analysis}

\begin{table}[t]
  \small
  \centering
  \makebox[3.1in][r]{
 \begin{tabular}{lcc}
    \hline
    \textbf{Feature}           & \multicolumn{2}{c}{\textbf{Language}} \\
                            & \textbf{Italian} & \textbf{Ladin}  \\
    \hline
    Average number of words per entry       & 65            & 70                           \\
    Average number of characters per entry  & 144           & 348                          \\
    Positive label count                       & 9,842         & 9,842                        \\
    Negative label count                       & 2,669         & 2,669                        \\
    \hline
  \end{tabular}
  }
  \caption{\label{SD-SA}
    Summary statistics of the synthetic paired sentiment analysis dataset in Italian and Ladin.
  }
\end{table}

To assess the qualitative aspects of translation performance using the FT-NLLB model, we analyze specific examples of translation errors from Italian to Ladin. This analysis helps identify common error types—such as lexical inaccuracies, syntactic issues, and morphological mismatches—and provides insight into the strengths and limitations of the machine-generated translations. For instance, consider the following sentence pair, where the Ladin translation is largely faithful to the Italian source, successfully preserving both structure and meaning.

\textbf{Italian}: \textit{Sono stata a un congresso a Torino e ho passato due notti al Best Western Crystal Palace, un hotel 4 stelle di fianco alla stazione di Porta Nuova. Ho apprezzato molto la vasca idromassaggio e il giornale ogni giorno: La Stampa, con un foglio di sintesi degli argomenti principali.}

\textbf{Ladin}: \textit{I sun stada a n congres a Torin y i à sté döes nöts tl Best Western Crystal Palace, n hotel da 4 stëres dlungia la staziun de Porta Nuova. I á dër aprijé la vasca da iade y le foliet da vigni de: La Stampa, cun na plata de ressumé di argomënc prinzipai.}

Upon closer inspection above, the Ladin translation only exhibits minor grammatical and lexical inaccuracies. The phrase “\textit{I á sté döes nöts}” should be corrected to “\textit{I sun stada döes nöts}” to ensure grammatical correctness. Additionally, “\textit{la vasca da iade}” ought to be revised to “\textit{la vasca idromassaje}” to more accurately reflect the original Italian phrase “\textit{la vasca idromassaggio}.” These examples illustrate the model’s general ability to produce semantically faithful translations, while also highlighting areas where linguistic precision can be improved. Overall, the model demonstrates strong potential for handling Ladin.

\subsection{Ladin synthetic datasets}

\begin{table}[t]
  \centering
  \small
  \makebox[3.2in][r]{
 \begin{tabular}{lcc}
    \hline
    \textbf{Feature}           & \multicolumn{2}{c}{\textbf{Language}} \\
                            & \textbf{Italian} & \textbf{Ladin}  \\
    \hline
    Average number of words per question       & 18    & 19   \\
    Average number of characters per question  & 108    & 97     \\
    Average number of words per choices        & 28    & 30  \\
    Average number of characters per choices   & 187    & 174  \\
    Frequency of entries with 3 choices         & 304    & 304   \\
    Frequency of entries with 4 choices         & 196    & 196   \\
    Frequency of entries with 5 choices         & 264    & 264    \\
    \hline
  \end{tabular}
  }
  \caption{\label{SD-MCQA}
    Statistics of the synthetic paired MCQA dataset in Italian and Ladin.
  }
\end{table}

\begin{table*}[t]
  \centering
  \small
  \makebox[\textwidth]{
    \begin{tabular}{lcccc@{\hspace{2em}}cccc}
      \hline
      \textbf{Model} &
      \multicolumn{4}{c}{\textbf{Ladin}} &
      \multicolumn{4}{c}{\textbf{Italian}} \\
      \cline{2-9}
      & \multicolumn{2}{c}{\textbf{SA}} & \multicolumn{2}{c}{\textbf{MCQA}}
      & \multicolumn{2}{c}{\textbf{SA}} & \multicolumn{2}{c}{\textbf{MCQA}} \\
      
      & \textbf{Acc} & \textbf{F1} & \textbf{Acc} & \textbf{F1}
      & \textbf{Acc} & \textbf{F1} & \textbf{Acc} & \textbf{F1} \\
      \hline
      FSL-LLaMA      & 93.99 & 96.07 & 44.20 & 44.46
                     & 97.92 & 98.18 & 54.79 & 54.18 \\
      m-DistilBERT   & 92.28 & 95.03 & 36.66 & 36.69  
                     & 94.78 & 96.41 & 30.73 & 30.29 \\
      XLM-RoBERTa    & 80.15 & 84.81 & 25.21 & 25.80  
                     & 97.14 & 98.08 & 26.14 & 26.14 \\
      mT5            & 50.09 & 69.37 & 23.50 & 23.16  
                     & 49.98 & 69.25 & 22.62 & 21.92 \\
      \hline
    \end{tabular}
  }
  \caption{\label{SA-MCQA} 
   Results for SA and MCQA tasks in Ladin, with Italian included for comparison.
  } 
\end{table*}

Since the FT-NLLB model demonstrates the best performance for translation in our experiments using \textit{AD\textsubscript{Ita\_Lad}} as the training data, we utilize this model to translate \textit{D\textsubscript{Ita}} into \textit{D\textsubscript{Lad}} through the framework shown in Figure~\ref{fig:experiments}.
Initially, we obtained 30,477 paired entries for (\textit{D\textsubscript{Ita\_SA}}, \textit{D\textsubscript{Lad\_SA}}) as the SA dataset after applying the first filtering step based on cosine similarity between Italian and Ladin with the threshold of 0.68. This value is derived from the cosine similarity between both languages in \textit{AD\textsubscript{Ita\_Lad}}. Subsequently, we perform back-translation from \textit{D\textsubscript{Lad\_SA}} to \( D^{\prime}\textsubscript{\textit{Ita\_SA}}\). The threshold for the second filtering step is determined based on the average (\(\mu\)) BLEU and METEOR scores calculated for each entry by comparing \textit{D\textsubscript{Ita\_SA}} with \( D^{\prime}\textsubscript{\textit{Ita\_SA}}\). These scores were found to be 33.63 and 0.58, respectively. The final filtered synthetic SA dataset, consisting of 12,511 entries, statistically is detailed in Table~\ref{SD-SA}. This SA dataset contains an average of 65 and 70 words per entry for Italian and Ladin, respectively. Additionally, the majority of entries in the final dataset—over 9,000 instances—are labeled with positive sentiment.

Concurrently, we obtained 4,651 paired entries of (\textit{D\textsubscript{Ita\_MCQA}}, \textit{D\textsubscript{Lad\_MCQA}}) as the MCQA dataset, following an initial filtering step based on cosine similarity, similar to the procedure used for the synthetic SA dataset. The thresholds for the second filtering step are also determined using the average BLEU and METEOR scores calculated between \textit{D\textsubscript{Ita\_MCQA}} and \( D^{\prime}\textsubscript{\textit{Ita\_MCQA}}\), which are 36.58 and 0.62, respectively. The final filtered synthetic MCQA dataset, comprising 764 entries, is presented in Table~\ref{SD-MCQA}, with the majority of questions featuring three answer choices. Notably, both Italian-Ladin paired datasets of SA and MCQA tasks exhibit a higher average number of words and characters per entry compared to the authentic parallel dataset \textit{AD\textsubscript{Ita\_Lad}}. 

\subsection{Synthetic data assessment for NLP tasks}

After obtaining the new synthetic datasets \textit{SD\textsubscript{Ita\_Lad}}
, we compare them against the manually translated golden dataset 
\( GD \) to evaluate translation quality. Specifically, we use cosine similarity metrics to evaluate the semantic alignment between the synthetic and manually created data. The cosine similarity scores of \textit{SD\textsubscript{Ita\_Lad}} and \( GD \), with respect to \textit{AD\textsubscript{Ita\_Lad}}, are 86.61 and 85.75, respectively. These results indicate a high degree of semantic consistency and minimal disparity between the synthetic and human-translated datasets.

Subsequently, we combine \textit{SD\textsubscript{Ita\_Lad}} with \textit{AD\textsubscript{Ita\_Lad}} to construct the training data for fine-tuning the FT-NLLB model as the MT model. The testing data \( T \) is then used to evaluate the performance of the resulting translation model on these combined datasets, as shown in the lower section of Table ~\ref{translation-performance}. Detailed results for the three subsets (\( t_1 \), \( t_2 \), and \( t_3 \)) of \( T \) are also presented in the lower section of Tables~\ref{add-trans_to_ladin} and~\ref{add-trans_to_italian}. We compare our results with the augmented synthetic translations generated by the RBMT system proposed by \citet{frontull24}, which incorporates additional monolingual data in both Ladin and Italian. As summarized in Table~\ref{translation-performance}, experimental results demonstrate that incorporating our synthetic data consistently improves performance across all evaluation metrics, surpassing the previous benchmark. In both translation directions, the FT-NLLB model performs well compared to its counterpart model.

Given that \textit{SD\textsubscript{Ita\_Lad}} encompasses SA and MCQA tasks in Ladin, we establish a benchmark to evaluate model performance on these tasks. We adopt the few-shot learning (FSL) using LLaMA 3.1 70B as our LLM's approach. Appendices~\ref{sec:fsl-sa} and~\ref{sec:fsl-mcqa} provide example prompts used in the FSL approach. In parallel, we assess the performance of several transformer-based models, including the Distilbert-base-multilingual-cased (m-DistilBERT)
model, the XLM-RoBERTa base model, and the mT5-small model. Table~\ref{SA-MCQA} summarizes the results for SA and MCQA tasks in Ladin, with Italian results included for comparison. The LLM-based approach achieves the highest scores in both tasks in terms of balanced accuracy (Acc) and F1-score (F1), although the improvements over other models are marginal. m-DistilBERT also performs well across both tasks. Specifically, the results indicate that the SA dataset in Ladin is generally well-handled by the evaluated models, with the exception of mT5. This indicates that our created Ladin SA dataset is sufficiently informative and well-aligned with the task objectives to support effective model training. In contrast, performance on the MCQA task remains low across all models, even when evaluated on Italian data. This suggests that the MCQA task is inherently more challenging than the SA task. The difficulty may be attributed to the limited size of the MCQA dataset and the broad topical coverage of the questions, both of which may hinder model generalization.

\section{Conclusions and Future Work}
In this work, we explore various strategies for translating Ladin—an extremely low-resource language—and demonstrate that our NLLB-based model consistently outperforms counterpart models. Beyond translation, we introduce the first comprehensive Ladin benchmark dataset covering machine translation, sentiment analysis, and multiple-choice question answering, all derived through synthetic augmentation from monolingual Italian resources. 

Our evaluation shows that while synthetic data enables competitive performance across tasks further progress is needed to enhance knowledge transfer from high-resource to low-resource languages. This study lays foundational work for future NLP research on Ladin and similarly underrepresented languages. 

Future work will investigate knowledge distillation techniques to more effectively transfer knowledge from high-resource languages to low-resource ones, as well as explore advanced methods to better exploit available monolingual Ladin data for improved model performance. We are also committed to further foster developments of Ladin resources and to engaging community in actively researching this language by establishing leaderboards based on the created datasets.

\section*{Limitations}

The Ladin-Italian dataset used in this study primarily consists of short sentences, which may limit the generalizability of the models to longer and more complex sentence structures. Moreover, the limited availability of manually crafted monolingual Italian MCQA datasets restricts the number of usable samples, which may impact the robustness and reliability of the experimental results. 



\bibliography{custom}

\begin{thebibliography}{39}
\providecommand{\natexlab}[1]{#1}

\bibitem[{Alabi et~al.(2022)Alabi, Adelani, Mosbach, and Klakow}]{Alabi:2022}
Jesujoba~O. Alabi, David~Ifeoluwa Adelani, Marius Mosbach, and Dietrich Klakow. 2022.
\newblock Adapting pre-trained language models to african languages via multilingual adaptive fine-tuning.
\newblock In \emph{Proceedings of the 29th International Conference on Computational Linguistics}, page 4336–4349.

\bibitem[{Bagheri~Nezhad et~al.(2025)Bagheri~Nezhad, Agrawal, and Pokharel}]{SinaBN}
Sina Bagheri~Nezhad, Ameeta Agrawal, and Rhitabrat Pokharel. 2025.
\newblock Beyond data quantity: Key factors driving performance in multilingual language models.
\newblock In \emph{Proceedings of the First Workshop on Language Models for Low-Resource Languages}, page 225–239.

\bibitem[{Bird(2020)}]{StevenBird}
Steven Bird. 2020.
\newblock \href {https://doi.org/10.18653/v1/2020.coling-main.313} {Decolonising speech and language technology}.
\newblock In \emph{Proceedings of the 28th International Conference on Computational Linguistics}, page 3504–3519.

\bibitem[{Cahyawijaya et~al.(2024)Cahyawijaya, Lovenia, and Fung}]{Cahyawijaya:2024}
Samuel Cahyawijaya, Holy Lovenia, and Pascale Fung. 2024.
\newblock \href {https://doi.org/10.18653/v1/2024.naacl-long.24} {Llms are few-shot in-context low-resource language learners}.
\newblock In \emph{Proceedings of the 2024 Conference of the North American Chapter of the Association for Computational Linguistics: Human Language Technologies}, volume~1, page 405–433.

\bibitem[{Cheng et~al.(2024)Cheng, Chen, Hu, Tang, Xu, and Ning}]{Cheng:24}
Qi~Cheng, Liqiong Chen, Zhixing Hu, Juan Tang, Qiang Xu, and Binbin Ning. 2024.
\newblock \href {https://doi.org/10.1016/j.nlp.2024.100099} {A novel prompting method for few-shot ner via llms}.
\newblock \emph{Natural Language Processing Journal}, 8(100099).

\bibitem[{Connor(2023)}]{connor2023ladin}
Anthony~Thomas Connor. 2023.
\newblock \emph{Ladin Perspectives on Language and Identity in the Central Dolomites of Northern Italy}.
\newblock Ph.d. thesis, The University of Sheffield, Faculty of Arts and Humanities, School of Languages and Cultures.

\bibitem[{Desole(2020)}]{desole2025}
Alessandro Desole. 2020.
\newblock \href {https://github.com/alessandrodesole/Sentiment-Analysis-on-Tripadvisor-reviews} {Sentiment analysis on tripadvisor reviews}.
\newblock Accessed: 2025-01-23.

\bibitem[{Erardi et~al.(2022)Erardi, Gardner, and Comploi}]{SilviaLadin}
Silvia Erardi, Ronald~L. Gardner, and Sara Comploi. 2022.
\newblock \href {https://doi.org/10.1177/00145858221107342} {Anglicisms in ladin: Loanwords and local perceptions}.
\newblock \emph{Forum Italicum}, 56(2):272–--306.

\bibitem[{Feng et~al.(2022)Feng, Yang, Cer, Arivazhagan, and Wang}]{labse:22}
Fangxiaoyu Feng, Yinfei Yang, Daniel Cer, Naveen Arivazhagan, and Wei Wang. 2022.
\newblock \href {https://doi.org/10.18653/v1/2024.americasnlp-1.22} {Language-agnostic bert sentence embedding}.
\newblock In \emph{Proceedings of the 60th Annual Meeting of the Association for Computational Linguistics}, volume~1, pages 878 -- 891.

\bibitem[{Frontull and Moser(2024)}]{frontull24}
Samuel Frontull and Georg Moser. 2024.
\newblock \href {https://doi.org/10.18653/v1/2024.loresmt-1.13} {Rule-based, neural and llm back-translation: Comparative insights from a variant of ladin}.
\newblock In \emph{Proceedings of the Seventh Workshop on Technologies for Machine Translation of Low-Resource Languages (LoResMT 2024)}, page 128–138.

\bibitem[{Gambardella et~al.(2024)Gambardella, Iwasawa, and Matsuo}]{Gambardella:2024}
Andrew Gambardella, Yusuke Iwasawa, and Yutaka Matsuo. 2024.
\newblock \href {https://doi.org/10.18653/v1/2024.acl-short.8} {Language models do hard arithmetic tasks easily and hardly do easy arithmetic tasks}.
\newblock In \emph{Proceedings of the 62nd Annual Meeting of the Association for Computational Linguistics}, volume~2, page 85–91.

\bibitem[{García(2024)}]{garcia:24}
Sofía García. 2024.
\newblock \href {https://doi.org/10.18653/v1/2024.wmt-1.84} {Enhaced apertium system: Translation into low-resource languages of spain spanish–asturian}.
\newblock In \emph{Proceedings of the Ninth Conference on Machine Translation}, page 878–884.

\bibitem[{Hatami et~al.(2024)Hatami, Banerjee, Arcan, Buitelaar, and McCrae}]{hatami:24}
Ali Hatami, Shubhanker Banerjee, Mihael Arcan, Paul Buitelaar, and John~Philip McCrae. 2024.
\newblock \href {https://doi.org/10.18653/v1/2024.wmt-1.76} {English-to-low-resource translation: A multimodal approach for hindi, malayalam, bengali, and hausa}.
\newblock In \emph{Proceedings of the Ninth Conference on Machine Translation}, page 815–822.

\bibitem[{Lim et~al.(2024)Lim, Yun, Kim, Choi, and Kim}]{lim:24}
Seong~Hoon Lim, Taejun Yun, Jinhyeon Kim, Jihun Choi, and Taeuk Kim. 2024.
\newblock Analysis of multi-source language training in cross-lingual transfer.
\newblock In \emph{Proceedings of the 62nd Annual Meeting of the Association for Computational Linguistics}, volume~1, page 712–725.

\bibitem[{Liu et~al.(2023)Liu, Li, and Wang}]{liu-multil:24}
Wuying Liu, Wei Li, and Lin Wang. 2023.
\newblock Multiloop incremental bootstrapping for lowresource machine translation.
\newblock In \emph{Proceedings of Machine Translation Summit XIX}, page 1–11.

\bibitem[{Lu et~al.(2025)Lu, Yang, Yang, Dong, Ma, Aihemaiti, Atawulla, Wang, and Zhou}]{DPO-LLM}
Kaiwen Lu, Yating Yang, Fengyi Yang, Rui Dong, Bo~Ma, Aihetamujiang Aihemaiti, Abibilla Atawulla, Lei Wang, and Xi~Zhou. 2025.
\newblock Low-resource language expansion and translation capacity enhancement for llm: A study on the uyghur.
\newblock In \emph{Proceedings of the 31st International Conference on Computational Linguistics}, page 8360–8373.

\bibitem[{Melchior(2023)}]{LucaM}
Luca Melchior. 2023.
\newblock \href {https://doi.org/10.1093/acrefore/9780199384655.013.723} {Raeto-romance: Romansh, ladin, friulian}.
\newblock \emph{Linguistics}.

\bibitem[{Mokhtarabadi et~al.(2025)Mokhtarabadi, Zamani, Maazallahi, and Hossein~Manshaei}]{FarsInstruct}
Hojjat Mokhtarabadi, Ziba Zamani, Abbas Maazallahi, and Mohammad Hossein~Manshaei. 2025.
\newblock Empowering persian llms for instruction following: A novel dataset and training approach.
\newblock In \emph{Proceedings of the First Workshop on Language Models for Low-Resource Languages}, page 31–67.

\bibitem[{Morim~da Silva et~al.(2024)Morim~da Silva, Srivastava, Ngoli, Röder, Moussallem, and Ngomo}]{morim2024}
Ana~Alexandra Morim~da Silva, Nikit Srivastava, Tatiana~Moteu Ngoli, Michael Röder, Diego Moussallem, and Axel-Cyrille~Ngonga Ngomo. 2024.
\newblock \href {https://doi.org/10.18653/v1/2024.loresmt-1.18} {Benchmarking low-resource machine translation systems}.
\newblock In \emph{Proceedings of the Seventh Workshop on Technologies for Machine Translation of Low-Resource Languages (LoResMT 2024)}, page 175–185.

\bibitem[{Murthy et~al.(2019)Murthy, Kunchukuttan, and Bhattacharyya}]{rudra:19}
Rudra Murthy, Anoop Kunchukuttan, and Pushpak Bhattacharyya. 2019.
\newblock \href {https://doi.org/10.18653/v1/N19-1387} {Addressing word-order divergence in multilingual neural machine translation for extremely low resource languages}.
\newblock In \emph{Proceedings of NAACL-HLT 2019}, page 3868–3873.

\bibitem[{Nguyen et~al.(2024)Nguyen, Aljunied, Joty, and Bing}]{Nguyen:2024}
Xuan-Phi Nguyen, Sharifah~Mahani Aljunied, Shafiq Joty, and Lidong Bing. 2024.
\newblock \href {https://doi.org/10.18653/v1/2024.acl-long.192} {Democratizing llms for low-resource languages by leveraging their english dominant abilities with linguistically-diverse prompts}.
\newblock In \emph{Proceedings of the 62nd Annual Meeting of the Association for Computational Linguistics}, volume~1, page 3501–3516.

\bibitem[{Pham et~al.(2024)Pham, Le, and Tuan}]{Pham:2024}
Trinh Pham, Khoi~M. Le, and Luu~Anh Tuan. 2024.
\newblock \href {https://doi.org/10.18653/v1/2024.acl-long.174} {Unibridge: A unified approach to cross-lingual transfer learning for low-resource languages}.
\newblock In \emph{Proceedings of the 62nd Annual Meeting of the Association for Computational Linguistics}, volume~1, page 3168–3184.

\bibitem[{Purason et~al.(2024)Purason, Kuulmets, and Fishel}]{purason2024llms}
Taido Purason, Hele-Andra Kuulmets, and Mark Fishel. 2024.
\newblock \href {https://arxiv.org/abs/2410.18902} {Llms for extremely low-resource finno-ugric languages}.
\newblock \emph{arXiv preprint arXiv:2410.18902}.
\newblock Cs.CL, version 1.

\bibitem[{Rajpoot et~al.(2024)Rajpoot, Bhat, and Shrivastava}]{rajpoot:24}
Pawan~Kumar Rajpoot, Nagraj~N Bhat, and Ashish Shrivastava. 2024.
\newblock \href {https://doi.org/10.18653/v1/2024.wmt-1.79} {Multimodal machine translation for low-resource indic languages: A chain-of-thought approach using large language models}.
\newblock In \emph{Proceedings of the Ninth Conference on Machine Translation}, page 833–838.

\bibitem[{Ranathunga et~al.(2023)Ranathunga, Lee, Skenduli, Shekhar, Alam, and Kaur}]{Ranathunga23}
Surangika Ranathunga, En-Shiun~Annie Lee, Marjana~Prifti Skenduli, Ravi Shekhar, Mehreen Alam, and Rishemjit Kaur. 2023.
\newblock \href {https://doi.org/10.1145/3567592} {Neural machine translation for low-resource languages: A survey}.
\newblock \emph{ACM Computing Surveys}, 55(11):1--37.

\bibitem[{Rinaldi et~al.(2024)Rinaldi, Gili, Francis, Goffetti, Patti, and Nissim}]{multi_it24}
Matteo Rinaldi, Jacopo Gili, Maria Francis, Mattia Goffetti, Viviana Patti, and Malvina Nissim. 2024.
\newblock Multiple choice questions on multiple topics in italian: A calamita challenge.
\newblock In \emph{CLiC-it 2024: Tenth Italian Conference on Computational Linguistics}.

\bibitem[{Shu et~al.(2024)Shu, Chen, Liu, Wang, Wu, Zhong, Li, Zhao, Jiang, Pan, Zhou, Owl, Zhai, Liu, Saunt, and Liu}]{RAG-LLM}
Peng Shu, Junhao Chen, Zhengliang Liu, Hui Wang, Zihao Wu, Tianyang Zhong, Yiwei Li, Huaqin Zhao, Hanqi Jiang, Yi~Pan, Yifan Zhou, Constance Owl, Xiaoming Zhai, Ninghao Liu, Claudio Saunt, and Tianming Liu. 2024.
\newblock \href {https://arxiv.org/abs/2411.11295} {Transcending language boundaries: Harnessing llms for low-resource language translation}.
\newblock \emph{arXiv preprint 2411.11295}.
\newblock Cs.CL, version 1.

\bibitem[{Su et~al.(2024)Su, Peng, Thillainathan, Guzmán, Ranathunga, and Lee}]{Su2024Unlocking}
Tong Su, Xin Peng, Sarubi Thillainathan, David Guzmán, Surangika Ranathunga, and En-Shiun~Annie Lee. 2024.
\newblock \href {https://doi.org/10.18653/v1/2024.findings-naacl.263} {Unlocking parameter-efficient fine-tuning for low-resource language translation}.
\newblock In \emph{Findings of the Association for Computational Linguistics: NAACL 2024}, page 4217–4225.

\bibitem[{Sánchez-Martínez et~al.(2024)Sánchez-Martínez, Pérez-Ortiz, Galiano-Jiménez, and Oliver}]{sanchez:24}
Felipe Sánchez-Martínez, Juan~Antonio Pérez-Ortiz, Aarón Galiano-Jiménez, and Antoni Oliver. 2024.
\newblock \href {https://doi.org/10.18653/v1/2024.wmt-1.57} {Findings of the wmt 2024 shared task translation into low-resource languages of spain: Blending rule-based and neural systems}.
\newblock In \emph{Proceedings of the Ninth Conference on Machine Translation}, page 684–698.

\bibitem[{Team et~al.(2022)Team, Costa-jussà, Cross, Çelebi, Elbayad, Heafield, Heffernan, Kalbassi, Lam, Licht, Maillard, Sun, Wang, Wenzek, Youngblood, Akula, Barrault, Gonzalez, Hansanti, Hoffman, Jarrett, Sadagopan, Rowe, Spruit, Tran, Andrews, Ayan, Bhosale, Edunov, Fan, Gao, Goswami, Guzmán, Koehn, Mourachko, Ropers, Saleem, Schwenk, and Wang}]{nllb2022}
NLLB Team, Marta~R. Costa-jussà, James Cross, Onur Çelebi, Maha Elbayad, Kenneth Heafield, Kevin Heffernan, Elahe Kalbassi, Janice Lam, Daniel Licht, Jean Maillard, Anna Sun, Skyler Wang, Guillaume Wenzek, Al~Youngblood, Bapi Akula, Loic Barrault, Gabriel~Mejia Gonzalez, Prangthip Hansanti, John Hoffman, Semarley Jarrett, Kaushik~Ram Sadagopan, Dirk Rowe, Shannon Spruit, Chau Tran, Pierre Andrews, Necip~Fazil Ayan, Shruti Bhosale, Sergey Edunov, Angela Fan, Cynthia Gao, Vedanuj Goswami, Francisco Guzmán, Philipp Koehn, Alexandre Mourachko, Christophe Ropers, Safiyyah Saleem, Holger Schwenk, and Jeff Wang. 2022.
\newblock \href {https://arxiv.org/abs/2207.04672} {No language left behind: Scaling human-centered machine translation}.
\newblock \emph{arXiv preprint arXiv:2207.04672}, v3.
\newblock Thu, 25 Aug 2022.

\bibitem[{Tran et~al.(2024)Tran, O’Sullivan, and Nguyen}]{tran24irish}
Khanh-Tung Tran, Barry O’Sullivan, and Hoang~D. Nguyen. 2024.
\newblock \href {https://doi.org/10.18653/v1/2024.loresmt-1.20} {Irish-based large language model with extreme low-resource settings in machine translation}.
\newblock In \emph{Proceedings of the Seventh Workshop on Technologies for Machine Translation of Low-Resource Languages (LoResMT 2024)}, page 193–202.

\bibitem[{Ul~Haq et~al.(2024)Ul~Haq, Huidrom, and Castilho}]{sami:dcu}
Sami Ul~Haq, Rudali Huidrom, and Sheila Castilho. 2024.
\newblock \href {https://doi.org/10.18653/v1/2024.wmt-1.75} {Dcu adapt at wmt24: English to low-resource multi-modal translation task}.
\newblock In \emph{Proceedings of the Ninth Conference on Machine Translation}, page 810–814.

\bibitem[{UNESCO(2010)}]{UNESCO}
UNESCO. 2010.
\newblock \emph{Atlas of the World’s Languages in Danger}.
\newblock UNESCO Publishing, Paris, France.

\bibitem[{Urbizu et~al.(2022)Urbizu, Vicente, Saralegi, Agerri, and Soroa}]{urbizu}
Gorka Urbizu, Inaki~San Vicente, Xabier Saralegi, Rodrigo Agerri, and Aitor Soroa. 2022.
\newblock Basqueglue: A natural language understanding benchmark for basque.
\newblock In \emph{Proceedings of the 13th Conference on Language Resources and Evaluation (LREC 2022)}, page 1603–1612.

\bibitem[{Valer et~al.(2024)Valer, Penzo, and Staiano}]{FasciaLadin}
Giovanni Valer, Nicolò Penzo, and Jacopo Staiano. 2024.
\newblock Nesciun lengaz lascià endò: Machine translation for fassa ladin.
\newblock In \emph{Proceedings of CLiC-it 2024: Tenth Italian Conference on Computational Linguistics}.

\bibitem[{Wang et~al.(2024)Wang, Wang, Arslan~Manzoor, Liu, Georgiev, Jyoti~Das, and Nakov}]{Yuxia:2024}
Yuxia Wang, Minghan Wang, Muhammad Arslan~Manzoor, Fei Liu, Georgi Georgiev, Rocktim Jyoti~Das, and Preslav Nakov. 2024.
\newblock \href {https://doi.org/10.18653/v1/2024.emnlp-main.1088} {Factuality of large language models: A survey}.
\newblock In \emph{Proceedings of the 2024 Conference on Empirical Methods in Natural Language Processing}, page 19519–19529.

\bibitem[{Yong et~al.(2024)Yong, Menghini, and Bach}]{yong2024lexc}
Zheng~Xin Yong, Cristina Menghini, and Stephen Bach. 2024.
\newblock \href {https://doi.org/10.18653/v1/2024.findings-emnlp.818} {Lexc-gen: Generating data for extremely low-resource languages with large language models and bilingual lexicons}.
\newblock In \emph{Findings of the Association for Computational Linguistics: EMNLP 2024}, pages 13990--14009.

\bibitem[{Zhang et~al.(2024)Zhang, Jijo, Setty, Chung, Javid, Vidra, and Clifford}]{zhang24llm}
Liang Zhang, Katherine Jijo, Spurthi Setty, Eden Chung, Fatima Javid, Natan Vidra, and Tommy Clifford. 2024.
\newblock \href {https://arxiv.org/abs/2402.01722} {Enhancing large language model performance to answer questions and extract information more accurately}.
\newblock \emph{arXiv preprint 2402.01722}.
\newblock Cs.CL, version 1.

\bibitem[{Zhang et~al.(2023)Zhang, Rajabi, Duh, and Koehn}]{zhang:mt23}
Xuan Zhang, Navid Rajabi, Kevin Duh, and Philipp Koehn. 2023.
\newblock \href {https://doi.org/10.18653/v1/2023.wmt-1.43} {Machine translation with large language models: Prompting, few-shot learning, and fine-tuning with qlora}.
\newblock In \emph{Proceedings of the Eighth Conference on Machine Translation}, page 468–481.

\end{thebibliography}

\appendix
\section{Additional Experimental Results}
\label{sec:add-results}
\begin{figure*}[tp]
  \centering
  \includegraphics[width=6.5in]{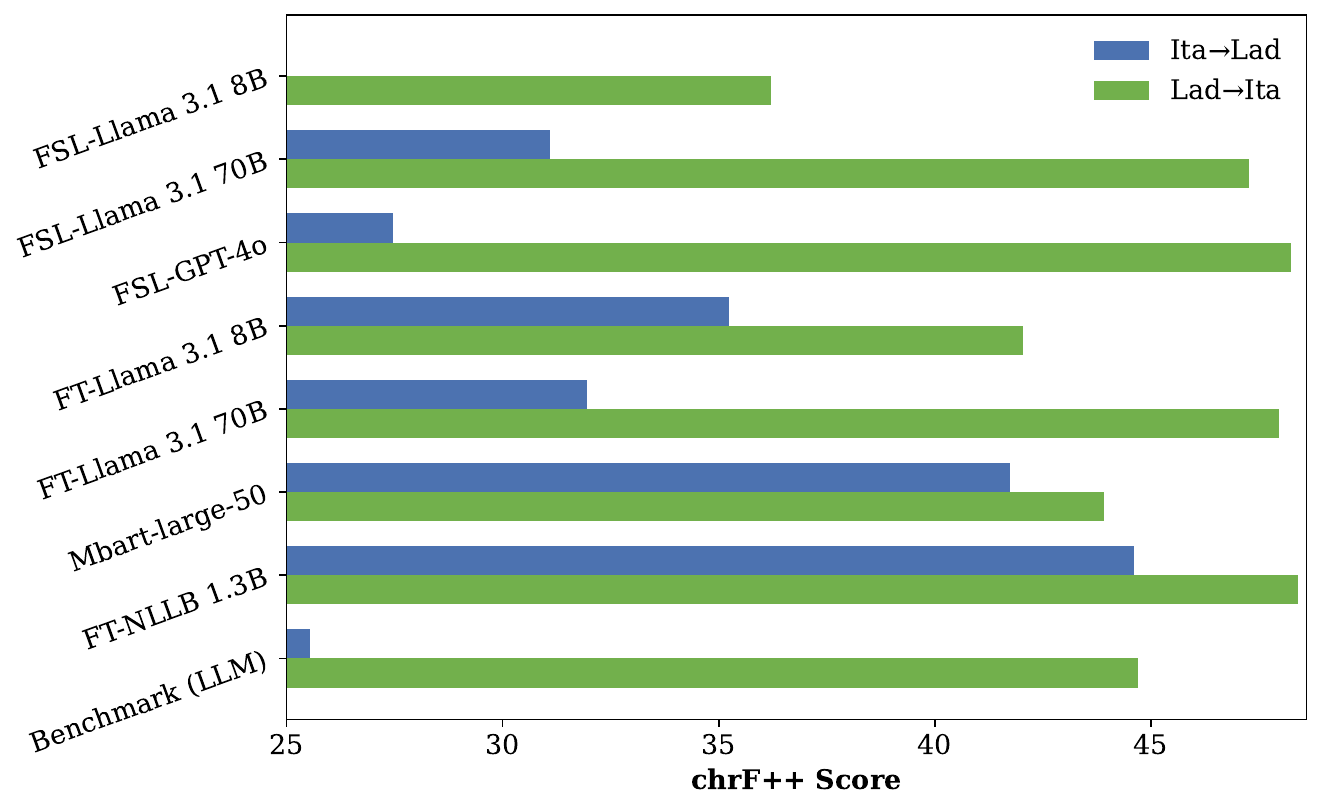}
  \caption{chrF++ scores of the models using \textit{AD\textsubscript{Ita\_Lad}} as the training data}
  \label{fig:chrf_AD}
\end{figure*}

\begin{figure*}[tp]
  \centering
  \includegraphics[width=6.5in]{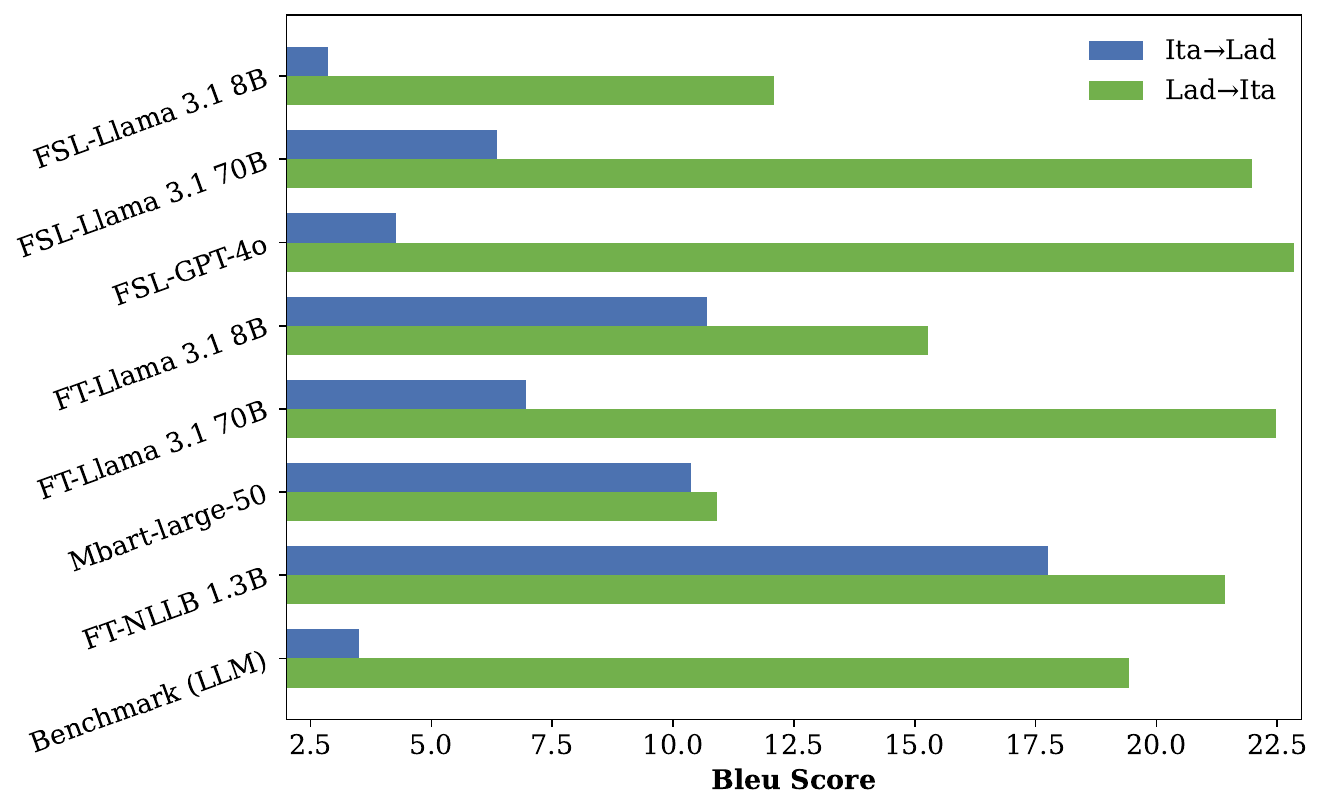}
  \caption{BLEU scores of the models using \textit{AD\textsubscript{Ita\_Lad}} as the training data}
  \label{fig:bleu_AD}
\end{figure*}

This section presents detailed experimental results for machine translation performance using \textit{AD\textsubscript{Ita\_Lad}} as the sole training data, as reported in the upper section of Table~\ref{translation-performance}. Figure~\ref{fig:chrf_AD} illustrates the chrF++ scores achieved by various models, while Figure~\ref{fig:bleu_AD} presents the corresponding BLEU score comparisons. 

\section{Back-translation Results}
As illustrated in Figure~\ref{fig:experiments}, we perform a second round of filtering by back-translating Ladin data into Italian and evaluating the results using BLEU and METEOR scores. For the SA dataset, which contains 30,477 samples, the average BLEU and METEOR scores are 33.63 and 0.58, respectively. In the case of the MCQA dataset (4,651 samples), the corresponding averages are 36.58 and 0.62. Below, we present examples comparing back-translated Italian samples with their original counterparts to facilitate a more detailed analysis.

\section{A Few-shot Learning Prompt Template in the MT task}
\label{sec:fsl}
\noindent\begin{verbatim}Here are examples of translations in a JSON 
format between Italian and Ladin with the 
Val Badia variant:
\end{verbatim}

\noindent\begin{verbatim}
{
    "translations": [
        {
        "Italian": "è venuta la mia ora",
        "Ladin": "al é gnü mia ora"
        },
        {
        "Italian": "vado dalle cugine!",
        "Ladin": "i vá dales jormanes"
        },
        {
        "Italian": "staccare la luce",
        "Ladin": "destodé la löm"
        },
        ...
        {
        "Italian": "a ottobre inoltrato",
        "Ladin": "d'otober fora"
        }
    ]
   }
\end{verbatim}

\noindent\begin{verbatim}Please provide the translation of the fol-
lowing 15 entries in the JSON format, fill-
ing the empty 'Ladin' fields for each entry.
Do not include any additional explanations 
or text:
\end{verbatim}


\noindent\begin{verbatim}
{
    "translations": [
        {
        "Italian": "imprimere nella mente",
        "Ladin": ""
        },
        {
        "Italian": "temperare la matita",
        "Ladin": ""
        },
        {
        "Italian": "mettere paura a qcn.",
        "Ladin": ""
        },
        ...
        {
        "Italian": "un animale scattante",
        "Ladin": ""
        }
    ]
   }
\end{verbatim}

\section{A Few-shot Learning Prompt Template in the SA task}
\label{sec:fsl-sa}
\noindent\begin{verbatim}Below are Tripadvisor reviews in Ladin (Val
Badia variant) along with their sentiment 
labels:
\end{verbatim}

\noindent\begin{verbatim}
    {
[review: "I sun stá chiló por 7 nes. Le gost
é é bun. Porimpó é i chelins cherdá indormed
í pormal. ...", label: 1], [review: "Lalber
ch é bunorté te na zona chîta dlungia na plaz
a. Gní zoruch vigni sëra ê sciöche ciafé nao
asa. ..." label: 0], 
...
[review: "I un passé chiló n bel fin dledema
hotel nët, personal da orëi bun, bun ince le
gosté y na posiziun ezelënta, impormó do lab
ela plaza San Marco. ...", label: 0]
    }

\end{verbatim}

\noindent\begin{verbatim}
Please classify the sentiment for the follo
wing 10 Tripadvisor reviews in Ladin (Val 
Badia variant) as either 0 (Positive) or 1
(Negative). Fill in the empty 'label' field
s with only 0 or 1. Respond with the sentime
nt labels in list format like this: [x, x, 
...]. Do not include any additional explana
tions or text.
\end{verbatim}


\noindent\begin{verbatim}
    {
[review: "Por na vistada y na fuga a Milan 
(na mostra, na cörta vijita, na spazirada)
él perfet. ...", label: ], [review: "Le bru
nch ne joav nia le prisc. Lhotel é n pü' dal
unc dal Duomo. ...", label: ], 
...
[review: "Rezeziun ezelënta, dantadöt le 
concierje cun sües racomanaziuns. ...", 
label: ]
    } 
\end{verbatim}

\section{A Few-shot Learning Prompt Template in the MCQA task}
\label{sec:fsl-mcqa}
\noindent\begin{verbatim}Below are multiple-choice questions in Lad
in (Val Badia variant) with 3, 4, or 5 answer 
choices. The correct answer is explicitly pr
ovided as an id number corresponding to the 
order of the choices:
\end{verbatim}

\noindent\begin{verbatim}
    {
[question: aladô dla lege 241/1990, olá é pa 
metüda sö la comisciun por l'azes ai documën
c aministratifs?, choices: ['pro la Presidën
za dl Consei di Minisć', 'Por vigni Entité pu
blica dl post', 'pro vigni Entité publica eco
nomica de competënza regionala'], answer: 0]
, [question: aladô dla DGR 514/2009, por PAI 
y PEI se intendi rispetivamënter:, choices: 
["Plan d'Assistënza Indicisé y Program Etich 
Individualisé", 'Program de Asistënza Indivi
duala y Program Educatif Individualisé', "Pl
an d' Assistënza Individualisé y Plan d'Educ
aziun Individualisé"], answer: 2],
...
[question: aladô de REICAT l'intestaziun uni
forme por na porsona:, choices: ['al corespo
gn tresala forma dl inom che vëgn dant tla pr
öma ediziundles operes dl autur.', 'al se ba
sa söl inom cun chël che la porsona medema é
generalmente identifiada.', 'ara ne pó mai e
ster metüda adöm da npseudonom.'], answer: 1]
    }
\end{verbatim}

\noindent\begin{verbatim}Please answer the questions based on the ava
ilable choices, by filling in the empty 'answ
er' fields with the id number corresponding to
the order of the choices. Provide the answers 
in a list format like this: [x, x, x, ..., x].
Do not include any additional explanations or
text. 
\end{verbatim}


\noindent\begin{verbatim}
    {
[question: na coleziun de figurines é metüda
adöm da 84 toc, 14 por vigni contignidú. 7 fi
gurines de vigni contignidú é lamincards, i 
restanc fotocartes. Tan de figurines de fo to
cartes ál pa la coleziun?, choices: ['42', '43'
, '44'], answer:], 
...
, [question: ci frasa, danter chëses listi
gades, á pa n complemënt de gauja?, choices:
['Laura é piada demez por n iade de plajëi',
'i ápormó cumpré na scincunda por le compliann
de Marco', "por na taiada de strom s'á ascensur
bloché"], answer: ]
   }
\end{verbatim}

\clearpage  

\end{document}